\documentclass[twoside,11pt]{article}

\usepackage{jmlr2e}
\setcitestyle{authoryear,aysep={,}}
\usepackage{blindtext}
\usepackage{amsmath,amssymb}
\usepackage{amsfonts}
\usepackage{amsthm}
\usepackage[font={small}]{caption}
\usepackage{listings}
\usepackage{xcolor}

\usepackage{lastpage}
\jmlrheading{25}{2024}{1-\pageref{LastPage}}{-/-; Revised -/-}{-/-}{21-0000}{Tommaso Carraro, Fabio Aiolli and Luciano Serafini}

\ShortHeadings{LTNtorch}{T. Carraro, L. Serafini and F. Aiolli}
\firstpageno{1}

\begin{document}

\title{LTNtorch: PyTorch Implementation of Logic Tensor Networks}

\author{\name Tommaso Carraro$^{1,2}$ \email tcarraro@fbk.eu \\
       \name Luciano Serafini$^2$ \email serafini@fbk.eu \\
       \name Fabio Aiolli$^1$ \email
       aiolli@math.unipd.it \\
       \addr $^1$ Department of Mathematichs, University of Padova, Padova, PD 35121, IT \\
       \addr $^2$ Data and Knowledge Management,
       Fondazione Bruno Kessler,
       Povo, TN 38123, IT
       }

\editor{My editor}

\maketitle

\begin{abstract}
Logic Tensor Networks (LTN) is a Neuro-Symbolic framework that effectively incorporates deep learning and logical reasoning. In particular, LTN allows defining a logical knowledge base and using it as the objective of a neural model. This makes learning by logical reasoning possible as the parameters of the model are optimized by minimizing a loss function composed of a set of logical formulas expressing facts about the learning task. The framework learns via gradient-descent optimization. Fuzzy logic, a relaxation of classical logic permitting continuous truth values in the interval $[0,1]$, makes this learning possible. Specifically, the training of an LTN consists of three steps. Firstly, $(i)$ the training data is used to ground the formulas. Then, $(ii)$ the formulas are evaluated, and the loss function is computed. Lastly, $(iii)$ the gradients are back-propagated through the logical computational graph, and the weights of the neural model are changed so the knowledge base is maximally satisfied. LTNtorch\footnote{Link to publicly available repository: \url{https://github.com/tommasocarraro/LTNtorch}.} is the fully documented and tested PyTorch implementation of Logic Tensor Networks. This paper presents the formalization of LTN and how LTNtorch implements it. Moreover, it provides a basic binary classification example.
\end{abstract}


\section{Introduction}

Neuro-Symbolic integration (NeSy)~\citep{nesy} is the branch of Artificial Intelligence that studies the incorporation of symbolic AI (e.g., logical reasoning) and sub-symbolic AI (e.g., deep learning). The idea of NeSy is to build systems that can merge the advantages of these two paradigms. For example, neural networks effectively learn in the presence of noise; however, they require a lot of samples and are usually black boxes. On the other hand, symbolic AI can perform few-shot learning and, in some cases, even zero-short learning. Plus, it is usually explainable by design as it is often implemented as a set of logical formulas. An effective integration could address the major limitations of neural networks. Logic Tensor Networks (LTN)~\citep{ltn} take a step in this direction by proposing a framework to learn neural networks using logical reasoning. In particular, LTN allows for the specification of logical loss functions and their minimization by gradient-descent optimization. A specific First-Order language, plus the use of fuzzy logic semantics~\citep{fuzzy} to implement logical operators, makes this learning possible. LTN was originally implemented in TensorFlow\footnote{First implementation of LTN: \url{https://github.com/logictensornetworks/logictensornetworks}.}. This paper presents the PyTorch implementation, dubbed LTNtorch. 

\section{Framework formalization}\label{sec:form}

To define the knowledge base, LTN uses a specific first-order language called Real Logic.
It is fully differentiable and has concrete semantics that allow mapping every symbolic expression into the domain of real numbers. This mapping function is referred to as \textit{grounding}
, denoted by $\mathcal{G}$. Thanks to $\mathcal{G}$, LTN can convert logical formulas into computational graphs that enable gradient-based optimization based on fuzzy logic semantics. Real Logic is a first-order~\citep{fol} language with a signature containing constant, variable, functional, and predicate symbols.
A term is constructed recursively from constants, variables, and functional symbols. An expression formed by applying a predicate symbol to some term(s) is called an atomic formula. Complex formulas are constructed recursively using connectives (i.e., $\lnot, \land, \lor, \implies, \leftrightarrow$) and quantifiers (i.e., $\forall, \exists$). 

\subsection{LTN grounding}

\begin{figure}[htp]
\includegraphics[width=\textwidth]{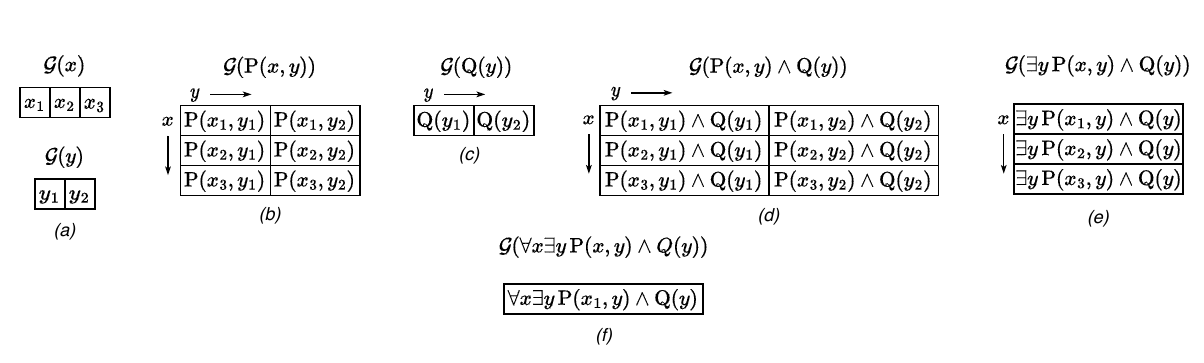}
\caption{Step by step computation of $\forall x \exists y \operatorname{P}(x,y) \land \operatorname{Q}(y)$ in LTN. $(a)$ Variables $x$ and $y$ are grounded as sequences of 3 and 2 individuals, respectively. Note each individual (e.g., $x_1$) is a real-valued tensor whose dimensions are determined by the dataset (e.g., image, vector, or scalar). $(b)$ Computation of binary predicate $\operatorname{P}$. Each item in this grid is a truth value in $[0,1]$. Note how LTN organizes the output in two dimensions, one for $x$ and one for $y$. $\operatorname{P}(x_1,y_1)$ is the evaluation of $\operatorname{P}$ on $x_1$ and $y_1$. $(c)$ Computation of unary predicate $\operatorname{Q}$, similar to $\operatorname{P}$. $(d)$ Computation of $\land$ applying $\mathcal{G}(\land)$ element-wise. Note that broadcasting is used to match the dimensions of $\mathcal{G}(\operatorname{P}(x,y))$ and $\mathcal{G}(\operatorname{Q}(y))$ before applying element-wise multiplication. $(e)$ Computation of $\exists$ by aggregating $y$'s dimension using $\mathcal{G}(\exists)$. $(f)$ Computation of $\forall$ by aggregating $x$'s dimension using $\mathcal{G}(\forall)$. \label{fig:grounding}}
\end{figure}

$\mathcal{G}$ maps each individual into a tensor of real features (e.g., sample's features), functions as real functions (e.g., regression functions), and predicates as real functions that specifically project onto a value in the interval $[0, 1]$ (e.g., classifiers). A variable $x$ is grounded to a \textit{sequence} of $n_x$ individuals from a domain (see Fig.~\ref{fig:grounding}$(a)$), with $n_x \in \mathbb{N}^{+}, n_x > 0$. As a consequence, a term $t(x)$ or a formula $\operatorname{P}(x)$, constructed recursively with variable $x$, will be grounded to a sequence of $n_x$ values too (see Fig.~\ref{fig:grounding}$(b,c)$). Afterward, connectives are grounded using fuzzy semantics
, while quantifiers using special aggregation functions. LTN provides different fuzzy semantics for these operators, namely \textit{Product}, \textit{Gödel}, and \textit{Lukasiewicz} semantics~\citep{fuzzy}. In what follows, we define the \textit{Product} configuration, better suited for gradient-based optimization~\citep{emilie}. In the formalization, $u,v,u_1,\dots,u_n \in [0,1]$, and $p \ge 1$. 

\small
\begin{gather*}
\mathcal{G}(\land) : u, v \mapsto u \cdot v \quad \mathcal{G}(\lor) : u, v \mapsto u + v - u \cdot v \quad
\mathcal{G}(\implies) : u, v \mapsto 1 - u + u * v \quad \mathcal{G}(\lnot) : u \mapsto 1 - u \\
\mathcal{G}(\exists) : u_1, \dots, u_n \mapsto (\frac{1}{n}\sum_{i=1}^{n} (u_i)^p)^{\frac{1}{p}} \quad
\mathcal{G}(\forall) : u_1, \dots, u_n \mapsto 1 - (\frac{1}{n}\sum_{i=1}^{n} (1 - u_i)^p)^{\frac{1}{p}}
\end{gather*}

\noindent 
Connective operators are applied element-wise to the tensors in input (see Fig.~\ref{fig:grounding}$(d)$), while aggregators aggregate the dimension of the tensor in input that corresponds to the quantified variable (see Fig.~\ref{fig:grounding}$(e,f)$). Figure~\ref{fig:grounding} gives an intuition of how LTN performs these operations to compute the formula $\forall x \exists y \operatorname{P}(x,y) \land \operatorname{Q}(y)$, where $\operatorname{P}$ and $\operatorname{Q}$ are logical predicates, while $x$ and $y$ are variables. Note these are also the steps LTNtorch uses to construct the computational graph for this formula.

\subsection{LTN learning}

Given a Real Logic knowledge base $\mathcal{K}=\{\phi_1,\dots,\phi_n\}$, where $\phi_1,\dots,\phi_n$ are closed formulas, LTN allows to learn the grounding of constants, functions, and predicates appearing in them.
In particular, if constants are grounded as embeddings, and functions/predicates onto neural networks, their grounding $\mathcal{G}$ depends on some learnable parameters $\boldsymbol{\theta}$. We denote a parametric grounding as $\mathcal{G(\cdot | \boldsymbol{\theta})}$. In LTN, the learning of parametric groundings is obtained by finding parameters $\boldsymbol{\theta}^*$ that maximize the satisfaction of $\mathcal{K}$, namely $\boldsymbol{\theta}^* = \operatorname{argmax}_{\boldsymbol{\theta}} \operatorname{SatAgg}_{\phi \in \mathcal{K}} \mathcal{G}(\phi | \boldsymbol{\theta})$. In the notation, $\operatorname{SatAgg} : [0,1]^* \mapsto [0, 1]$ is a formula aggregating operator, often defined using $\mathcal{G}(\forall)$. It represents the overall satisfaction of the knowledge base $\mathcal{K}$. As the objective is to maximize the overall satisfaction of $\mathcal{K}$, LTNtorch minimizes $1. - \operatorname{SatAgg}_{\phi \in \mathcal{K}} \mathcal{G}(\phi | \boldsymbol{\theta})$. Because Real Logic grounds expressions in continuous domains, LTN attaches gradients to every sub-expression and consequently learns through gradient-descent optimization.

\section{Binary classification in LTNtorch}\label{sec:ex}
We introduce a simple binary classification task
where the objective is to discriminate between cats and dogs. The building blocks for this example are $(1)$ a variable $dog$ to iterate through batches of dog images, $(2)$ a variable $cat$ to iterate through batches of cat images, and $(3)$ a unary predicate $\operatorname{Dog}$ distinguishing between cats and dogs. 
Given these symbols, the knowledge base LTNtorch maximizes to solve this task is $\mathcal{K} = \{\phi_1, \phi_2\}$, where $\phi_1 = \forall dog \operatorname{Dog}(dog)$ and $\phi_2 = \forall cat \lnot\operatorname{Dog}(cat)$. When $\mathcal{K}$ is maximally satisfied, $\operatorname{Dog}$ returns high truth values when dogs are given and low truth values (see the use of $\lnot$ in $\phi_2$) when cats are given.

\begin{figure}[htp]
\includegraphics[width=\textwidth]{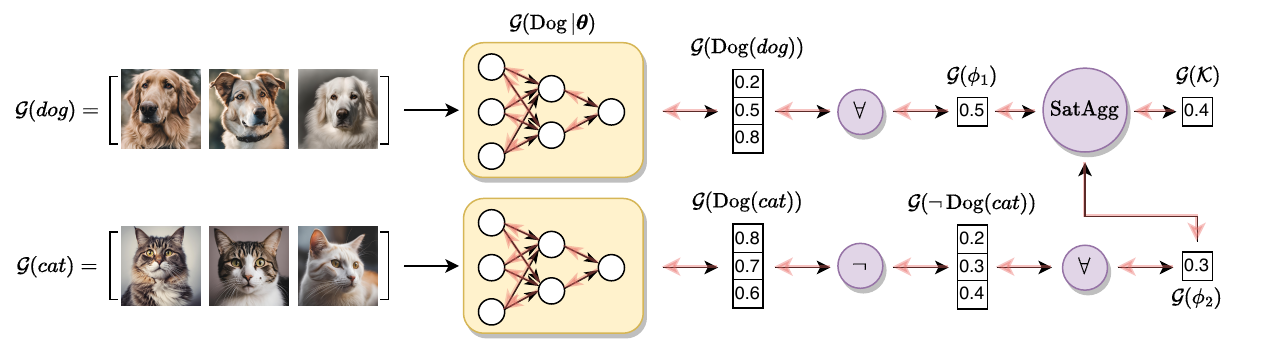}
\caption{Computational graph of the LTN for binary classification. Batch size for $\mathcal{B}_{dog}$ and $\mathcal{B}_{cat}$ is 3. Red arrows indicate how gradients are back-propagated through the logical structure to change the weights of predicate $\operatorname{Dog}$ so $\mathcal{K}$ is maximally satisfied. LTNtorch minimizes $1. - \mathcal{G}(\mathcal{K}) \equiv \mathcal{L}(\boldsymbol{\theta})$.\label{fig:comp}}
\end{figure}

LTNtorch makes use of $\mathcal{G}$ to build the computational graph for this task. In particular, $\mathcal{G}(dog) = \mathbb{R}^{b \times l \times w \times c}$, where $b$ is the batch dimension, and $l$, $w$, and $c$ are the length, width, and number of channels of the image, respectively. In other words, $dog$ is grounded as a batch of dog images. Similarly, $cat$ is grounded as a batch of cat images (see Fig.~\ref{fig:comp}). Then, $\mathcal{G}(\operatorname{Dog}|\boldsymbol{\theta}) : \mathbf{x} \mapsto \sigma (\operatorname{CNN}_{\boldsymbol{\theta}}(\mathbf{x}))$, namely $\operatorname{Dog}$ is implemented through a Convolutional Neural Network (i.e., CNN) with parameters $\boldsymbol{\theta}$. Specifically, this predicate takes images in input and returns truth values indicating how plausible it is that the images depict dogs (see Fig.~\ref{fig:comp}). Note the logistic function $\sigma$ allows $\operatorname{Dog}$ to be interpreted as a fuzzy logic predicate, as the output is in the interval $[0,1]$. The loss function for this task is: 

$$\mathcal{L}(\boldsymbol{\theta}) = 1. - \operatorname{SatAgg}_{\phi \in \mathcal{K}} \mathcal{G}_{\substack{dog\leftarrow \mathcal{B}_{dog}\\cat \leftarrow \mathcal{B}_{cat}}}(\phi | \boldsymbol{\theta})$$

\noindent where $\mathcal{K}=\{\phi_1,\phi_2\}$ and the notation $\substack{dog\leftarrow \mathcal{B}_{dog}}$ (similarly $\substack{cat\leftarrow \mathcal{B}_{cat}}$) means variable $dog$ is grounded with images from $\mathcal{B}_{dog}$, which is a batch of dog images randomly sampled from the set of dog images in the dataset. Figure~\ref{fig:comp} shows how LTNtorch builds the computational graph for this basic task. 

\subsection{LTNtorch implementation}

In what follows, we provide a partial code snippet implementing the training loop of the presented example in LTNtorch. We invite the reader to review the examples in the repository and the framework documentation to complete this implementation. Note the implementation details of \texttt{CNN\_model} and \texttt{train\_data\_loader} are not given as it is standard PyTorch code.

\begin{lstlisting}[language=Python, caption=LTNtorch implementation of the dog classification example., basicstyle=\tiny]
import ltn
# define the Dog predicate
Dog = ltn.Predicate(CNN_model())
# define logical operators and formula aggregator (i.e., SatAgg)
Not = ltn.Connective(ltn.fuzzy_ops.NotStandard())
Forall = ltn.Quantifier(ltn.fuzzy_ops.AggregPMeanError(p=2), quantifier="f")
SatAgg = ltn.fuzzy_ops.SatAgg()
# training loop
for epoch in range(n_epochs):
    train_loss = 0.0
    for i, (dog_imgs, cat_imgs) in enumerate(train_data_loader):
        optimizer.zero_grad()
        # ground logical variables with current training batch
        dog = ltn.Variable("dog", dog_imgs) # positive examples
        cats = ltn.Variable("cats", cat_imgs) # negative examples
        # compute loss function
        sat_agg = SatAgg(
            Forall(dogs, Dog(dogs)),  # this is phi_1
            Forall(cats, Not(Dog(cats)))  # this is phi_2
        )
        loss = 1. - sat_agg
        # back-propagation
        loss.backward()
        optimizer.step()
        train_loss += loss.item()
    train_loss = train_loss / len(train_data_loader)
\end{lstlisting}

\section{Conclusions}
In this paper, we introduced LTNtorch, the fully documented and tested PyTorch implementation of Logic Tensor Networks. We summarized the formalization of the framework and provided a simple example to help understand how LTNtorch implements the LTN. This paper does not have to be considered a comprehensive framework documentation. We strongly invite practitioners interested in approaching LTNs to read the original paper and the documentation carefully.








\vskip 0.2in
\bibliography{sample}

\begin{thebibliography}{5}
\providecommand{\natexlab}[1]{#1}
\providecommand{\url}[1]{\texttt{#1}}
\expandafter\ifx\csname urlstyle\endcsname\relax
  \providecommand{\doi}[1]{doi: #1}\else
  \providecommand{\doi}{doi: \begingroup \urlstyle{rm}\Url}\fi

\bibitem[Badreddine et~al.(2022)Badreddine, {d'Avila Garcez}, Serafini, and Spranger]{ltn}
Samy Badreddine, Artur {d'Avila Garcez}, Luciano Serafini, and Michael Spranger.
\newblock Logic tensor networks.
\newblock \emph{Artificial Intelligence}, 303:\penalty0 103649, 2022.
\newblock ISSN 0004-3702.
\newblock \doi{https://doi.org/10.1016/j.artint.2021.103649}.
\newblock URL \url{https://www.sciencedirect.com/science/article/pii/S0004370221002009}.

\bibitem[Cintula et~al.(2011)Cintula, H{\'a}jek, Noguera, et~al.]{fuzzy}
Petr Cintula, Petr H{\'a}jek, Carles Noguera, et~al.
\newblock \emph{Handbook of Mathematical Fuzzy Logic. Volume 1}, volume~37.
\newblock College Publications, 2011.

\bibitem[d’Avila Garcez et~al.(2009)d’Avila Garcez, Lamb, and Gabbay]{nesy}
Artur~S. d’Avila Garcez, Luís~C. Lamb, and Dov~M. Gabbay.
\newblock \emph{Neural-Symbolic Learning Systems}, pages 35--54.
\newblock Springer Berlin Heidelberg, Berlin, Heidelberg, 2009.
\newblock ISBN 978-3-540-73246-4.
\newblock URL \url{https://doi.org/10.1007/978-3-540-73246-4_4}.

\bibitem[Flach(2007)]{fol}
Peter~A. Flach.
\newblock \emph{First-Order Logic}, pages 35--68.
\newblock Springer Berlin Heidelberg, Berlin, Heidelberg, 2007.
\newblock ISBN 978-3-540-74113-8.
\newblock \doi{10.1007/978-3-540-74113-8_2}.
\newblock URL \url{https://doi.org/10.1007/978-3-540-74113-8_2}.

\bibitem[{van Krieken} et~al.(2022){van Krieken}, Acar, and {van Harmelen}]{emilie}
Emile {van Krieken}, Erman Acar, and Frank {van Harmelen}.
\newblock Analyzing differentiable fuzzy logic operators.
\newblock \emph{Artificial Intelligence}, 302:\penalty0 103602, 2022.
\newblock ISSN 0004-3702.
\newblock \doi{https://doi.org/10.1016/j.artint.2021.103602}.
\newblock URL \url{https://www.sciencedirect.com/science/article/pii/S0004370221001533}.

\end{thebibliography}

\end{document}